\documentclass{article}

    \PassOptionsToPackage{numbers}{natbib}

    \usepackage[final]{neurips_2023}

\usepackage[utf8]{inputenc} 
\usepackage[T1]{fontenc}    
\usepackage{hyperref}       
\usepackage{url}            
\usepackage{booktabs}       
\usepackage{amsfonts}       
\usepackage{nicefrac}       
\usepackage{microtype}      
\usepackage{xcolor}         

\usepackage{amsmath}
\usepackage{amsthm}
\usepackage{algorithm}
\usepackage{algorithmic}
\usepackage{enumitem}
\usepackage{graphicx}
\usepackage{caption}
\usepackage{wrapfig}
\usepackage{amssymb}
\newcommand{\bx}{\mathbf{x}}

\newcommand{\RE}{\ensuremath{\mathbb{R}}}

\newtheorem{theorem}{Theorem} 
\newtheorem{proposition}[theorem]{Proposition}

\theoremstyle{definition}  

\newtheorem*{definition*}{Definition}
\newtheorem*{remark*}{Remark}

\title{Generalized Contrastive Divergence: Joint Training \\of Energy-Based Model and Diffusion Model \\through Inverse Reinforcement Learning}

\author{%
  Sangwoong Yoon$^1$,  Dohyun Kwon$^2$, Himchan Hwang$^3$, Yung-Kyun Noh$^{1,4}$, Frank C. Park$^{3,5}$  \\
$^1$Korea Institute for Advanced Study,
$^2$University of Seoul,\\
$^3$Seoul National University,
$^4$Hanyang University,
$^5$Saige Research\\
\texttt{swyoon@kias.re.kr}, 
\texttt{dhkwon@uos.ac.kr}, \\
\texttt{himchan@robotics.snu.ac.kr}, 
\texttt{nohyung@hanyang.ac.kr}, 
\texttt{fcp@snu.ac.kr}\\
}

\begin{document}

\maketitle

\begin{abstract}
We present Generalized Contrastive Divergence (GCD), a novel objective function for training an energy-based model (EBM) and a sampler simultaneously.
GCD generalizes Contrastive Divergence \citep{hinton2002training}, a celebrated algorithm for training EBM, by replacing Markov Chain Monte Carlo (MCMC) distribution with a trainable sampler, such as a diffusion model.
In GCD, the joint training of EBM and a diffusion model is formulated as a minimax problem, which reaches an equilibrium when both models converge to the data distribution.
The minimax learning with GCD bears interesting equivalence to inverse reinforcement learning, where the energy corresponds to a negative reward, the diffusion model is a policy, and the real data is expert demonstrations.
We present preliminary yet promising results showing that joint training is beneficial for both EBM and a diffusion model.
GCD enables EBM training without MCMC while improving the sample quality of a diffusion model.
\end{abstract}

\section{Introduction}
Diffusion models \cite{ho2020ddpm,sohl-dickstein15} have achieved tremendous success in generating realistic high-dimensional samples. However, several challenges are remaining. 
First, a diffusion model does not directly minimize the deviation between the model and the data distributions. While recent theories show that diffusion models indirectly minimize a divergence between the model and data \cite{song2021maximum,kwon2022score,chen2023sampling}, a direct minimization can be more effective, if possible \cite{fan2023optimizing}.
Second, generation is slow, requiring a large number of denoising steps. 
Shortening the number of steps often incurs a trade-off in sample quality \cite{fan2023optimizing,kong2021fast}.
Third, the marginal likelihood of data can not be directly computed, and only its lower bound or importance-weighted approximation is available \cite{ho2020ddpm,sohl-dickstein15}.

In this paper, we propose Generalized Contrastive Divergence (GCD), a novel objective function motivated by Contrastive Divergence \cite{hinton2002training}, that shows training a diffusion model in conjunction with an energy-based model (EBM) helps tackle these problems.
First, learning with GCD is equivalent to the minimization of integral probability metric (IPM; e.g., Wasserstein distance) between a diffusion model and data, where the energy works as a critic in IPM.
Second, therefore, a diffusion model with fewer steps can be fine-tuned with GCD to improve sample quality.
Third, the jointly trained EBM can compute the marginal likelihood of data (up to constant).

This paper presents preliminary experimental results showing that GCD learning works successfully on a synthetic dataset and can be effective in improving the sample quality of the DDPM sampler, especially when the number of steps is small. We expect to obtain experimental results on larger real-world datasets in the near future. Furthermore, we discuss the connection between GCD learning and inverse reinforcement learning. This connection opens up potential opportunities for advancing generative modeling by leveraging the recent ideas of reinforcement learning.

The contributions of this paper can be summarized as follows:
\begin{itemize}[leftmargin=1em,topsep=0pt,noitemsep]
    \item We propose GCD learning, a novel method for jointly training EBM and a parametric sampler. The joint training is beneficial for both models.
    \begin{itemize}
        \item GCD learning enables EBM training without MCMC, which is often computationally expensive and unstable.
        \item GCD learning can be used to train a diffusion-style sampler with a small number of timesteps.
    \end{itemize}
    \item We show that GCD learning has interesting equivalence to the integral probability metric minimization problems (e.g., WGAN) and inverse reinforcement learning.
\end{itemize}

\section{Preliminaries}

\textbf{Diffusion for Sampling.} In this paper, we focus on discrete-time diffusion-based samplers, such as DDPM \cite{ho2020ddpm}. A sampler first draws a Gaussian noise vector $\bx_0$ and then iteratively samples from conditional distributions $\bx_{t+1}\sim \pi(\bx_{t+1}|\bx_{t},t)$ to obtain a final sample $\bx_T$. We will denote the final samples as $\bx=\bx_T$ and their marginal distribution as $\pi(\bx)$.
The initial samples $\bx_0$ are drawn from $\mathcal{N}(0,I)$.
The sampler $\pi(\bx)$ may be pre-trained through existing diffusion model techniques.



\textbf{Energy-Based Models.}
An energy-based model (EBM) $q(\bx)$ uses a scalar function called an energy $E(\bx)$ to represent a probability distribution:
\begin{align}
    q(\bx) = \frac{1}{Z}\exp(-E(\bx)/\tau), \quad E: \mathcal{X}\to\RE \quad Z=\int_{\mathcal{X}} \exp(-E(\bx)/\tau) d\bx, \label{eq:ebm}
\end{align}
where $\tau >0$ is temperature, $\mathcal{X}$ is the compact domain of data, and the normalization constant $Z$ is assumed to be finite.
The integral is replaced with a summation for discrete input data.

While multiple training algorithms for EBM have been proposed (see \cite{song2021train} for an overview), the standard method is maximum likelihood training which is equivalent to minimization of Kullback-Leibler (KL) divergence between data and a model: $\min_{q \in \mathcal{Q}} KL(p || q)$, where $p(\bx)$ is the data distribution, and $\mathcal{Q}$ being the set of feasible $q(\bx)$'s. KL divergence is defined as $KL(p || q)=\int_{\mathcal{X}} p(\bx) \log (p(\bx)/q(\bx)) d\bx$. Meanwhile, the exact computation of the log-likelihood gradient requires convergent MCMC sampling.

\textbf{Contrastive Divergence \cite{hinton2002training}.}
Hinton proposed an alternative objective function for training EBM which does not require convergent MCMC.
Let us write $\mathcal{T}_{q}$ as a single-step MCMC operator designed to draw sample from $q(\bx)$ and $\mathcal{T}_{q}^{k}(p)$ as the distribution of points after $k$ steps of MCMC transition $\mathcal{T}_q$ starting from $p(\bx)$.
Then, Contrastive Divergence (CD) learning is defined as follows:
\begin{align}
    \min_{q\in \mathcal{Q}} KL(p||q) - KL(\mathcal{T}_{q}^{k}(p)||q). \label{eq:cd}
\end{align}
CD learning is known to converge when $KL(p||q) \geq KL(\mathcal{T}_{q}^{k}(p)||q)$ is guaranteed \citep{lyu2011Unifying}.
However, CD still requires convergent MCMC when drawing a new sample from $q(\bx)$.

\section{Generalized Contrastive Divergence}

\subsection{Generalized Contrastive Divergence Learning}

We present Generalized Contrastive Divergence (GCD), an objective function for the joint training of EBM and a sampler.
From CD (Eq. \ref{eq:cd}), we replace MCMC distribution $\mathcal{T}_{q}^{k}(p)(\bx)$ with an arbitrary sampler $\pi(\bx)$ which will later be set as a DDPM-like discrete-time diffusion model. We also introduce maximization with respect to $\pi(\bx)$, motivated by the fact that $\mathcal{T}_{q}^{k}(p)(\bx)$ converges towards $q(\bx)$.
\begin{definition*}[GCD]
Suppose the data distribution $p(\bx)$, the EBM $q(\bx)$, and the sampler $\pi(\bx)$ are supported on the domain $\mathcal{X}$. Let $\mathcal{Q}$ and $\Pi$ be the sets of feasible $q(\bx)$'s and $\pi(\bx)$'s, respectively. \textbf{GCD learning} is defined as the following minimax problem:
\begin{align}
    \inf_{q\in\mathcal{Q}} \textcolor{blue}{\sup_{\pi\in\Pi}} KL(p||q) - KL(\textcolor{blue}{\pi} ||q), \label{eq:gcd}
\end{align}    
\end{definition*}
where the changes made from CD are highlighted. 
This minimax problem has a desirable equilibrium.
\begin{proposition}[Equilibrium]
    Suppose $p(\bx)\in\mathcal{Q}=\Pi$. Then, $p(\bx)=q(\bx)=\pi(\bx)$ is the equilibrium of GCD learning.
\end{proposition}
Eq. \ref{eq:gcd} can be rewritten with respect to energy. Plugging EBM's definition (Eq.~\ref{eq:ebm}), we obtain the following equivalent problem, which we solve in practice:
\begin{align}
\label{eq:gcd-learning-energy}
\inf_{E\in\mathcal{E}}\sup_{\pi\in\Pi}  \mathbb{E}_{p}[E(\bx)]  - \mathbb{E}_{\pi}[E(\bx)] + \tau \mathcal{H}(\pi),
\end{align}
where $\mathcal{H}(\pi)=-\int \pi(\bx) \log \pi(\bx)d\bx $ is the differential entropy of $\pi(\bx)$. The set of energy functions $\mathcal{E}$ is derived from $\mathcal{Q}$:\\
\begin{align}
    \mathcal{E} := \left\{ E : \mathcal{X} \rightarrow \mathbb{R} | \exp(-E(\bx)/\tau)/Z \in \mathcal{Q} \hbox{ where } Z = \int \exp(-E(\bx)/\tau) d\bx < \infty \right\}.
\end{align}
In other words, $E \in \mathcal{E}$ if and only if $E(\bx) = - \tau \log q(\bx) + c$ for some $q \in \mathcal{Q}$ and a constant $c$.
Similarly to Eq. \ref{eq:gcd}, GCD learning in energy (Eq. \ref{eq:gcd-learning-energy}) also reaches the equilibrium when $E(\bx)=-\tau \log p(\bx) + c$ and $\pi(\bx)=p(\bx)$ (see Appendix \ref{app:proof_prop2}).
GCD learning has interesting theoretical connections to other generative modeling problems and inverse reinforcement learning. 

\subsection{Equivalence to Entropy-Regularized Integral Probability Metric Minimization}
Generative modeling is often formulated as the minimization of Integral Probability Metric (IPM; \cite{ipm}), which measures the deviation between the data distribution and the model. WGAN \cite{arjovsky17wasserstein} is a well-known example.
GCD learning (Eq. \ref{eq:gcd-learning-energy}) can also be viewed as the minimization of IPM between $p(\bx)$ and $\pi(\bx)$ but under entropy regularization for $\pi(\bx)$.
The energy $E(\bx)$ works as a critic in IPM, and the feasible set $\mathcal{E}$ characterizes IPM (hence we write IPM as  $D^\mathcal{E}(p||\pi)$).

\begin{proposition}[Entropy-regularized IPM minimization] \label{prop:ipm}
    Assume that $p(\bx)\in \mathcal{Q} \cap \Pi$. In addition, suppose that $\mathcal{E}$ is closed under negation. Consider the following problem:
    \begin{align}
        \inf_{\pi\in\Pi} D^\mathcal{E}(p||\pi) - \tau \mathcal{H}(\pi), 
        \hbox{ where } D^\mathcal{E}(p||\pi) := \sup_{E\in \mathcal{E}} \lvert\mathbb{E}_{p}[E(\bx)] - \mathbb{E}_{\pi}[E(\bx)]\rvert. \label{eq:entropy-reg-ipm-min}
    \end{align}
    This problem has the same optimal value and the same equilibrium point ($E(\bx)=-\tau \log p(\bx) + c$ and $\pi(\bx)=p(\bx)$) to the GCD learning in energy (Eq. \ref{eq:gcd-learning-energy}).   
\end{proposition}
Proposition \ref{prop:ipm} holds because Eq. \ref{eq:gcd-learning-energy} and Eq. \ref{eq:entropy-reg-ipm-min} are the primal and the dual problems of the same objective function, where total duality holds.
A detailed proof is in Appendix \ref{app:proof_prop2}.

In conventional IPM minimization problems, such as WGAN, where $\tau=0$ and $\mathcal{E}$ being the set of 1-Lipschitz functions, the critic is not guaranteed to converge to a quantity related to $\log p(\bx)$.
The entropy regularization is critical for obtaining an accurate energy estimate and is the key difference from a prior work \cite{fan2023optimizing}, where DDPM is optimized for only IPM.



\subsection{Equivalence to Maximum Entropy Inverse Reinforcement Learning}
GCD learning can also be interpreted through the lens of reinforcement learning (RL).
Eq. \ref{eq:gcd-learning-energy} is a special case of maximum entropy inverse reinforcement learning (IRL; \cite{ziebart2008maximum,ziebart2010modeling,ho2016generative}).
The connection between diffusion modeling and Markov Diffusion Processes (MDP) is well noted in previous works \cite{fan2023optimizing,fan2023reinforcement}.
The sampler $\pi(\bx)$ is an agent with a stochastic policy $\pi(\bx_{t+1}|\bx_{t},t)$.
The intermediate samples $\bx_0,\bx_1,\ldots,\bx_T(=\bx)$, combined with time indices, form a trajectory of states $\{(\bx_t, t)\}_{t=0}^{T}$.
The action corresponds to the choice of the next state.

However, as no explicit reward signal is available in generative modeling, the reward must be inferred from the data, making the problem one of IRL.
A training data $\bx$ serves as expert demonstrations for the terminal state $\bx_T$.
There is no demonstration for intermediate states, and thus the reward is only given for the terminal state.
Then, a natural choice of the reward for a sampler is $\log p(\bx)$, which is unknown.
In GCD learning, the energy function is responsible for learning $\log p(\bx)$, providing the reward signal $-E(\bx)$ for training $\pi(\bx)$.

One notable features of GCD is that it maximizes the terminal state entropy $\mathcal{H}(\pi(\bx_T))$ instead of the causal entropy $\mathcal{H}(\pi(\bx_{t+1}|\bx_t,t))$, as is done in other max-ent IRL methods \cite{ziebart2008maximum,ziebart2010modeling,ho2016generative}.
The connection to IRL allows us to analyze the training from RL perspective and to employ techniques proven to be effective in RL.

\section{Joint Training of EBM and Diffusion Model}
\label{sec:training}
Now we present an algorithm for GCD learning. While GCD learning is not restricted on the choice of $\pi(\bx)$, we shall focus on the case where $\pi(\bx)$ is a DDPM-style sampler.
We write $\theta$ and $\phi$ as the parameters of EBM $q_\theta(\bx)$ and a sampler $\pi_\phi(\bx)$, respectively. GCD learning can be written as follows:
\begin{align}
    \min_{\theta} \max_{\phi} \mathcal{L}(\theta, \phi), \; \mathcal{L}=\mathbb{E}_{p}[E_\theta(\bx)]  - \mathbb{E}_{\pi_\phi}[E_\theta(\bx)] + \tau \mathcal{H}(\pi_\phi).
\end{align}
To solve this minimax problem, we alternatively update the EBM $E_\theta(\bx)$ and the diffusion model $\pi_\phi(\bx)$ as typically done in gradient descent-ascent algorithms, e.g., \cite{lin20gda}. Temperature $\tau$ is treated as a hyperparameter.

\begin{figure}[t]
    \centering
    \includegraphics[width=0.99\textwidth]{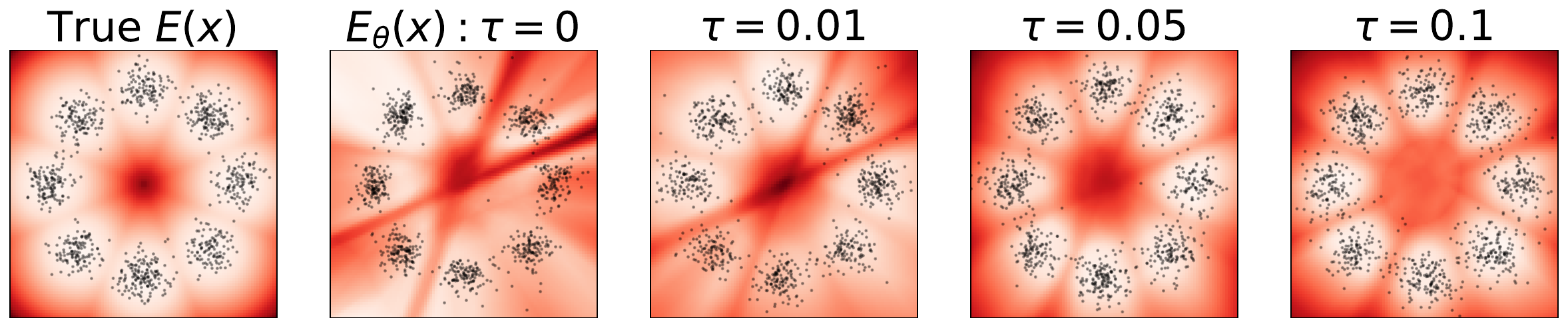}
    \caption{2D Density Estimation on 8 Gaussians. The red shades indicate the energy (white is low), and the dots are samples.  Without entropy regularization ($\tau=0$), the energy does not reflect the data distribution, and the samples are collapsed to the mode.}
    \label{fig:2d-experiment}

    \centering
    \includegraphics[width=0.98\textwidth]{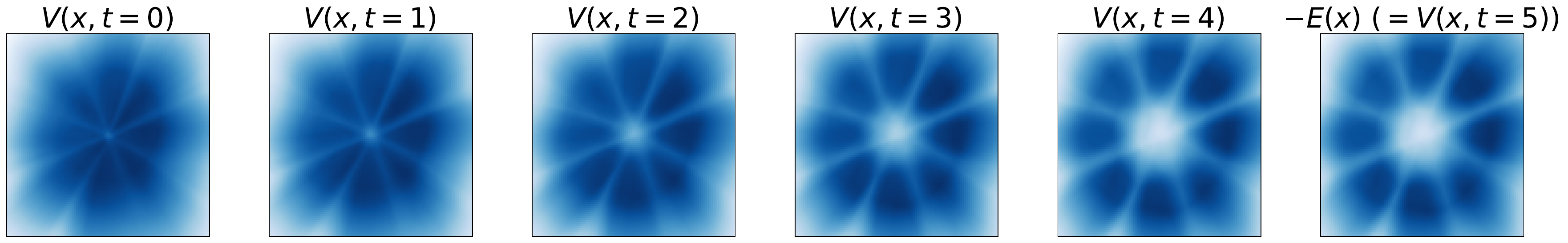}
    \caption{Value functions at each time step ($\tau=0.1$ case). Blue indicates a high value. }
    \label{fig:values}
\end{figure}

\textbf{EBM Update.}
The energy is updated to discriminate the training data and the samples from $\pi_\phi(\bx)$.
\begin{align}
    \nabla_\theta \mathcal{L} =& \mathbb{E}_{p(\bx)}[\nabla_\theta E_\theta(\bx)] - \mathbb{E}_{\pi_\phi(\bx)}[\nabla_\theta E_\theta(\bx)]. \label{eq:ebm_update_on} 
\end{align}
This update corresponds to the reward learning step in IRL \cite{ziebart2008maximum,ho2016generative}.

\textbf{Diffusion Model Update.}
The sampler is updated to maximize the reward $-E_\theta(\bx)$ under entropy regularization.
When a sampler admits a reparametrization trick, we may directly take the gradient of $-E_\theta(\bx)$. However, the reparametrization trick may not be applicable to a diffusion model due to the memory overhead. Instead, we can employ REINFORCE-style policy gradient algorithm \cite{williams1992simple} (Derivation in Appendix \ref{app:pg-derivation}):
\begin{align}
    \nabla_\phi \mathcal{L} &= \nabla_\phi \left(\mathbb{E}_{\pi_\phi(\bx)}[-E_\theta(\bx)]+\tau\cdot\mathcal{H}(\pi_\phi) \right) = \mathbb{E}_{\pi_\phi(\bx)}\left[ \nabla_\phi \log \pi_\phi(\bx)( -E_\theta(\bx)-\tau\log \pi_\phi(\bx))\right] \nonumber\\
    &= \mathbb{E}_{\pi_\phi(\bx_{0:T})} \left[ \sum_{t=0}^{t={T-1}} \nabla_\phi \log \pi_\phi(\bx_{t+1}|\bx_t,t) \left[-E_\theta(\bx_T) - \tau\log \pi_\phi(\bx_T) - B(\bx_t,t)\right]  \right], \label{eq:pg-entropy-baseline}
\end{align}
where the term $-\tau\log \pi_\phi(\bx_T)$ can be seen as an additional reward for exploration. This update corresponds to the forward RL in typical IRL. 
For sample efficiency, we use proximal policy optimization \cite{schulman2017proximal} to update the parameters multiple times for training data.
The baseline function $B(\bx_t,t)$ is introduced to reduce the variance \cite{greensmith2004variance} and is described shortly.

\textbf{Marginal Probability Estimation.}
The log probability $\log \pi_\phi(\bx_T)$ in Eq. \ref{eq:pg-entropy-baseline}  can not be directly computed in DDPM.
We estimate $\log \pi_\phi(\bx_T)$ using mini-batch samples from $\pi_\phi(\bx)$.
Following \cite{du2021improved}, we employ the nearest-neighbor density estimator \cite{kozachenko1987sample}, which gives $\log \pi_\phi(\bx) \approx D\log d_{k}(\bx)+c$ where $D$ is the dimensionality of data, $d_{k}(\bx)$ is the distance to $k$-nearest neighbor, and $c$ is constant with respect to $\bx$.

\textbf{Baseline Estimation.}
We model the baseline of $-E_\theta(\bx)$ and $-\tau \log \pi_\phi(\bx_T)$ separately. For energy, we employ the state value function as the baseline $V(\bx_t,t)=\mathbb{E}_{\pi_\phi}[-E_\theta(\bx_T)|\bx_t]$.
We parametrize $V(\bx,t)$ as a separate model and estimate its parameter by minimizing temporal difference loss $\min_{V(\cdot,t)} (V(\bx_{t+1}, t+1) - V(\bx_t, t))^2$ for $t=0,\ldots,T-1$.
Meanwhile, we simply use the estimated entropy as the baseline $\tau\mathcal{H}(\pi_\phi)=-\tau\mathbb{E}_{\pi_\phi}[\log\pi_\phi(\bx_T)]$, where the expectation is approximated using exponential moving average. As a result, the baseline is given as $B(\bx_t,t) = V(\bx_t, t) + \tau \mathcal{H}(\pi_\phi)$.




\begin{wraptable}{r}{0.5\textwidth}
\setlength\tabcolsep{3pt}
    \small
    \begin{tabular}{lcccc}
    \toprule
         Method & T & $\tau$ & $SW$ ($\downarrow$) & AUC ($\uparrow$) \\
    \midrule
         DDPM &5&-& 0.967$\pm$0.005 & -\\
         DDPM &10&-& 0.824$\pm$0.002 & -\\
         DDPM &100&-& 0.241$\pm$0.003 & -\\
         DDPM &1000&-& 0.123$\pm$0.014 & - \\
         GCD-Scratch &5&0.05& 0.114$\pm$0.025 & 0.867\\
         GCD-FT &5 &0.05& \textbf{0.086}$\pm$0.008 & \textbf{0.880}\\
         GCD-FT &5&0& 0.152$\pm$0.008 & 0.513 \\
    \bottomrule
    \end{tabular}
    \vskip 0.1in
    \caption{Fine-tuning DDPM for 8 Gaussians. $SW$ indicates sliced Wasserstein-2 distance (1,000 projections) between samples and data estimated from 10k samples. The standard deviation is computed from 5 independent samplings.
    AUC indicates how well the energy discriminates the uniform distribution over the domain from data. AUC is computed with 10k samples. The maximum value of AUC vs the uniform distribution is about 0.9059.}
    \label{tab:2d-finetune}
\end{wraptable}

\section{Experiments}

In our experiment, we show the synergistic benefit of jointly training EBM and a diffusion model on 2D 8 Gaussians data.
We use time-conditioned MLP networks for both a value network and a policy network (i.e., sampler). The time step is encoded into a 128D vector using sinusoidal positional embedding and concatenated into the hidden neurons of MLP. The last time step ($t=T$) of the value network is treated as the negative energy. 
The conditional distributions $\pi(\bx_{t+1}|\bx_t,t)$ in the sampler are Gaussians where the mean is determined by the policy network. We use the linear variance schedule from $10^{-5}$ to $10^{-2}$. Throughout the experiment, we use $T=5$ for GCD.

We first train both EBM and a sampler from scratch (Figure \ref{fig:2d-experiment}).
In the conventional IPM minimization ($\tau=0$), the critic, i.e., the energy, deviates significantly from the true energy of $p(\bx)$. Furthermore, the samples tend to collapse to the mode of the data density, under-representing the variance of data. The entropy regularization makes the energy produce $q(\bx)$ that accurately reflects $p(\bx)$. Meanwhile, the value functions reflect the evolution of the energy from the initial Gaussian density (Figure \ref{fig:values}), similar to Diffusion Recovery Likelihood \cite{gao2021learning}. 

Next, we show GCD can improve the sample quality of pre-trained DDPM samplers (Table \ref{tab:2d-finetune}). DDPM samplers with varying $T$ are trained on 8 Gaussians until convergence, and Wasserstein distances from their samples to real data are measured. The quality of samples deteriorates significantly as $T$ becomes small. However, if we further fine-tune DDPM using GCD learning (GCD-FT), we obtain a Wasserstein distance even smaller than that of $T=1000$ DDPM. This result is possible because GCD learning directly minimizes IPM between the samples and data. Similar results were reported in \cite{fan2023optimizing}. If we do not fine-tune the sampler (GCD-Scratch), the distance slightly increases, showing there is a gain from DDPM pretraining. 
It is interesting to note that GCD learning with entropy regularization ($\tau = 0.05$), outperforms conventional IPM minimization ($\tau = 0$).
One hypothesis is that the entropy term facilitates the exploration and thus leads to a better policy.



\section{Related Work}
GCD is an attempt to train generative samplers using RL. When applying RL, a key design choice is defining the reward signal. 
One source of reward is human feedback \cite{clark2023directly,zhang2023hive}, which became popular after being applied in large language models \cite{ouyang2022}. 
However, since human feedback is typically costly, a machine learning model, such as an image aesthetic quality estimator, can be used as a substitute for the reward function \cite{fan2023reinforcement,black2023training}. Assuming the absence of human feedback or an external reward function, GCD employs the IRL approach where the reward function is inferred from demonstrations. 
In GCD, the reward is the log-likelihood of data, learned from training data through EBM.
The previous work of Fan and Lee \cite{fan2023optimizing} can also be seen as IRL, but their reward function is not interpreted as the log-likelihood.

GCD can be seen as a novel method for EBM training that does not require MCMC. EBMs are a powerful class of generative models that shows promising performance in compositional generation \cite{du2020compositional} and out-of-distribution detection \cite{yoon2021autoencoding,yoon2023energybased}. However, EBM suffers from instability in training, often requiring computationally expensive MCMC that is difficult to tune and not convergent in practice \cite{du2019,du2021improved,xiao2021vaebm,gao2021learning,lee2023guiding}.
GCD enables the stable training of EBM by removing MCMC from training.

GCD bears a formal resemblance to algorithms that train EBM using minimax formulation. 
CoopFlow \cite{xie2022flow} jointly trains a normalizing flow sampler and EBM, but the flow is not directly optimized to minimize the divergence to EBM.
Divergence Triangle \cite{Han_2019_CVPR} trains a sampler, an encoder, and EBM simultaneously, resulting in a more complex problem than a minimax problem.
Due to the entropy regularization, GCD may have a connection to entropy-constrained
optimal transport problems \cite{cuturi2013sinkhorn,mokrov2023energy} which regularizes the entropy of the joint distribution. However, the entropy of a single marginal distribution is regularized in GCD.

\section{Conclusion}
Bridging generative modeling and IRL, GCD opens up new opportunities to improve generative modeling by leveraging tools from reinforcement learning.
We are planning to scale up the experiments to large-scale tasks, such as image generation.
Besides generation tasks, GCD may also be useful for other tasks that require an accurate energy estimation, such as out-of-distribution detection \cite{du2021improved,yoon2021autoencoding}.

\begin{ack}
S. Yoon was supported by a KIAS Individual Grant (AP095701) via the Center for AI and Natural Sciences at Korea Institute for Advanced Study.
Y.-K. Noh was partly supported by NRF/MSIT (No. 2018R1A5A7059549, 2021M3E5D2A01019545) and IITP/MSIT (IITP-2021-0-02068, 2020-0-01373, RS-2023-00220628).
This work was supported in part by 
IITP-MSIT grant 2021-0-02068 (SNU AI Innovation Hub),
IITP-MSIT grant 2022-0-00480 (Training and Inference Methods for Goal-Oriented AI Agents),  
KIAT grant P0020536 (HRD Program for Industrial Innovation), 
ATC+ MOTIE Technology Innovation Program grant 20008547, 
SRRC NRF grant RS-2023-00208052, 
SNU-AIIS, 
SNU-IAMD, 
SNU BK21+ Program in Mechanical Engineering, 
and SNU Institute for Engineering Research.
\end{ack}

\bibliography{ref}
\bibliographystyle{unsrt}

\clearpage
\appendix

\section{Proofs}

\subsection{Proof of Proposition 1} \label{app:proof_prop1}

Solving the inner maximization yields $\pi^*(\bx)=q(\bx)$, which is attainable since $\Pi=\mathcal{Q}$. Consequently, the outer minimization simplifies to $\min_{q\in\mathcal{Q}} KL(p||q)$, which reaches its minimum at $q^*(\bx)=p(\bx)$, attainable as $p \in \mathcal{Q}$.



\subsection{Proof of Proposition 2} \label{app:proof_prop2}

We will prove Proposition 2 by showing that GCD learning in energy (Eq. \ref{eq:gcd-learning-energy}) and the entropy-regularized IPM minimization (Eq. \ref{eq:entropy-reg-ipm-min}) are the primal and the dual problems of the same objective function, where total duality holds.
Total duality implies strong duality, and thus the primal and the dual optimal values are the same.

\paragraph{Primal and Dual Problems.}

We consider a function $\Phi: \mathcal{E}\times\Pi \longrightarrow \mathbb{R}$ defined as follows: 
\begin{align}
\Phi(E, \pi) = \mathbb{E}_{p}[E(\bx)]  - \mathbb{E}_{\pi}[E(\bx)] + \tau\;\mathcal{H}(\pi).
\label{eq:obj_function}
\end{align}
The \textbf{primal} problem generated by $\Phi(E, \pi)$ is given as
\begin{align}
    \inf_{E \in \mathcal{E}}\sup_{\pi \in \Pi} \mathbb{E}_{p}[E(\bx)]  - \mathbb{E}_{\pi}[E(\bx)] + \tau\;\mathcal{H}(\pi),
    \label{eq:primal}
\end{align}
which is equivalent to GCD learning in energy (Eq. \ref{eq:gcd-learning-energy}).
The \textbf{dual} problem generated by $\Phi(E, \pi)$ is equivalent to the minimization of IPM under entropy regularization:
\begin{align}
    \inf_{\pi \in \Pi}\sup_{E \in \mathcal{E}} -\mathbb{E}_{p}[E(\bx)]  + \mathbb{E}_{\pi}[E(\bx)] - \tau\;\mathcal{H}(\pi) 
    &=\inf_{\pi \in \Pi}\sup_{E\in \mathcal{E}} \lvert\mathbb{E}_{p}[E(\bx)] - \mathbb{E}_{\pi}[E(\bx)]\rvert - \tau \; \mathcal{H}(\pi) \label{eq:dual_prob-2}  \\
    &= \inf_{\pi\in\Pi} D^\mathcal{E}(p||\pi) - \tau\; \mathcal{H}(\pi),
\end{align}
where we used the condition that $\mathcal{E}$ is closed under negation to obtain Eq. \ref{eq:dual_prob-2}.

\paragraph{Total Duality.}
We will show that total duality holds between these primal and dual problems. 
The function $\Phi$ is linear (and hence convex) for $E(\bx)$ and concave for $\pi(\bx)$.
Then, total duality holds if and only if $\Phi$ has a saddle point (\cite[Ch.\ 1, pp. 12-14]{ryu2022large}).

Let us define a function $E^p(\bx)= -\tau\log{p(\bx)} + c$, where $c$ being a constant, which 
 can be considered as the energy of $p(\bx)$ computed under the temperature $\tau$. 
Let us show that $(E^*, \pi^*)=(E^p, p)$ is the saddle point of $\Phi$ for any constant $c$ in $E^p$. Note that $(E^p, p) \in \mathcal{E} \times \Pi$ as $p \in \mathcal {Q} \cap \Pi$.
\begin{align}
    \Phi(E^p, \pi) &= \mathbb{E}_{p}[E^p(\bx)] + \mathbb{E}_{\pi}[\tau\log p - \tau\log \pi]\\
    &= \mathbb{E}_{p}[E^p(\bx)] -\tau KL(\pi || p) \leq \Phi(E^p,p) \quad \forall \pi \in \Pi \label{eq:duality-left}
\end{align}
Also,
\begin{align}
    \Phi(E, p) &= \mathbb{E}_{p}[E(\bx)]  - \mathbb{E}_{p}[E(\bx)] + \tau\mathcal{H}(p)\\
    &= \tau\mathcal{H}(p) = \Phi(E^p, p) \quad \forall E \in \mathcal{E} \label{eq:duality-right}
\end{align} 
Combining Eq. \ref{eq:duality-left} and Eq. \ref{eq:duality-right}, we obtain $\Phi(E^p, \pi)\leq \Phi(E^p, p) \leq \Phi(E, p)$ for all $(E, \pi) \in \mathcal{E}\times\Pi$, and thus $(E^p, p)$ is a saddle point of $\Phi$.  
The saddle point is the equilibrium for both the primal and the dual problems.

\section{Algorithm}

\begin{algorithm}[H]
   \caption{Generalized Contrastive Divergence }
   \label{alg:GCD}
\begin{algorithmic}
   \STATE Initialize sample buffer $\mathcal{B}=\varnothing$. 
   \STATE Set $i=0$. Set sampler update period $n_{sampler}$ and the number of PPO update $n_{PPO}$.
   \FOR{ $\bx$ in dataset}
   \STATE Obtain $m$ samples from $\pi(\bx)$ and append them into $\mathcal{B}$. \\
   \STATE Update EBM $E(\bx)$ using $\mathcal{B}$ by Eq. \ref{eq:ebm_update_on}. \\
   \STATE Update value functions $V(\bx_t, t)$'s using $\mathcal{B}$. \\
   \IF{$i \;\%\; n_{sampler} = 0$}
   \FOR{$j=1,...,n_{PPO}$}
   \STATE Update $\pi(\bx)$ using PPO using $\mathcal{B}_{\text{on}}$. \\
   \ENDFOR
   \STATE Reset $\mathcal{B}$ as $\varnothing$.
   \ENDIF
   \STATE $i \leftarrow i +1$
   \ENDFOR
\end{algorithmic}
\end{algorithm}


\section{Derivation of Policy Gradient}
\label{app:pg-derivation}

Here, we provide a detailed derivation on the policy gradient update for sample parameters (Eq. \ref{eq:pg-entropy-baseline}).
\begin{align}
    \nabla_\phi \mathcal{L} 
    &= \nabla_\phi \int \pi_\phi(\bx_T) [-E_\theta(\bx_T) - \tau\log \pi_\phi(\bx_T) ] d\bx_T \\
    &= \int \nabla_\phi \pi_\phi(\bx_T) [-E_\theta(\bx_T) - \tau\log \pi_\phi(\bx_T)] d\bx_T \\
    &= \int \nabla_\phi \left(\int  \pi_\phi(\bx_{0:T}) d\bx_{0:{T-1}} \right) [-E_\theta(\bx_T) - \tau\log \pi_\phi(\bx_T)] d\bx_T \\
    &= \int \nabla_\phi\pi_\phi(\bx_{0:T}) [-E_\theta(\bx_T) - \tau\log \pi_\phi(\bx_T)]d\bx_{0:T}\\
    &= \int \pi_\phi(\bx_{0:T}) \nabla_\phi \log \pi_\phi(\bx_{0:T}) [-E_\theta(\bx_T) - \tau\log \pi_\phi(\bx_T) ] d\bx_{0:T} \\
    &= \mathbb{E}_{\pi_\phi(\bx_{0:T})} \left[ \nabla_\phi \log \pi_\phi(\bx_{0:T})[-E_\theta(\bx_T)-\tau\log\pi_\phi(\bx_T)] \right] \\
    &=\mathbb{E}_{\pi_\phi(\bx_{0:T})} \left[ \left(\sum_{t=0}^{t={T-1}} \nabla_\phi \log \pi_\phi(\bx_{t+1}|\bx_t,t)\right)[-E_\theta(\bx_T)-\tau\log\pi_\phi(\bx_T)]  \right].
\end{align}


\end{document}